\documentclass[conference]{IEEEtran}
\IEEEoverridecommandlockouts
\usepackage{cite}
\usepackage{amsmath,amssymb,amsfonts}
\usepackage{algorithmic}
\usepackage{graphicx}
\usepackage{textcomp}
\usepackage{comment}
\usepackage{xcolor}
\usepackage{soul}
\usepackage{subcaption}
\usepackage{float}
\def\BibTeX{{\rm B\kern-.05em{\sc i\kern-.025em b}\kern-.08em
    T\kern-.1667em\lower.7ex\hbox{E}\kern-.125emX}}
\begin{document}

\title{Intelligent Dynamic Handover via AI-assisted Signal Quality Prediction in 6G Multi-RAT Networks

\thanks{This work was supported in part by UNITY-6G Project, funded by the European Union’s HORIZON-JU-SNS-2024 program, under Grant Agreement No 101192650;
and in part by the 6G-Cloud Project funded from the European Union’s HORIZON-JU-SNS-2023 program under Grant Agreement No 101139073.}
}

\author{
\IEEEauthorblockN{Maria Lamprini A. Bartsioka}
    \IEEEauthorblockA{\textit{R\&D Department}\\
    \textit{Four Dot Infinity} \\
    Athens, Greece \\
    mbarts@fourdotinfinity.com}
\and
\IEEEauthorblockN{Anastasios Giannopoulos}
    \IEEEauthorblockA{\textit{R\&D Department}\\
    \textit{Four Dot Infinity} \\
    Athens, Greece \\
    angianno@fourdotinfinity.com}
\and
\IEEEauthorblockN{Sotirios Spantideas}
    \IEEEauthorblockA{\textit{R\&D Department}\\
    \textit{Four Dot Infinity} \\
    Athens, Greece \\
    sospanti@fourdotinfinity.com}
}

\maketitle

\begin{abstract}
The emerging paradigm of 6G multiple Radio Access Technology (multi-RAT) networks, where cellular and Wireless Fidelity (WiFi) transmitters coexist, requires mobility decisions that remain reliable under fast channel dynamics, interference, and heterogeneous coverage. Handover in multi-RAT deployments is still highly reactive and event-triggered, relying on instantaneous measurements and threshold events. This work proposes a Machine Learning (ML)-assisted Predictive Conditional Handover (P-CHO) framework based on a model-driven and short-horizon signal quality forecasts. We present a generalized P-CHO sequence workflow orchestrated by a RAT Steering Controller, which standardizes data collection, parallel per-RAT predictions, decision logic with hysteresis-based conditions, and CHO execution. Considering a realistic multi-RAT environment, we train RAT-aware Long Short Term Memory (LSTM) networks to forecast the signal quality indicators of mobile users along randomized trajectories. The proposed P-CHO models are trained and evaluated under different channel models for cellular and IEEE 802.11 WiFi integrated coverage. We study the impact of hyperparameter tuning of LSTM models under different system settings, and compare direct multi-step versus recursive P-CHO variants. Comparisons against baseline predictors are also carried out. Finally, the proposed P-CHO is tested under soft and hard handover settings, showing that hysteresis-enabled P-CHO scheme is able to reduce handover failures and ping-pong events. Overall, the proposed P-CHO framework can enable accurate, low-latency, and proactive handovers suitable for ML-assisted handover steering in 6G multi-RAT deployments.
\end{abstract}


\section{Introduction}

Fifth-generation (5G) and beyond (B5G) communication networks are being deployed worldwide to address the growing demands for massive device connectivity, high capacity, ultra-low latency, and seamless communication across diverse scenarios \cite{spantideas2024smart}. A key architectural approach to meet these requirements is the deployment of heterogeneous networks (HetNets). HetNets can be classified into two main categories: (i) those integrating multiple Radio Access Technologies (multi-RATs), and (ii) those comprising heterogeneous equipment types. Such architectures have been progressively incorporated into Third-Generation Partnership Project (3GPP) specifications starting from Release 11 \cite{3GPP_TR_36839} and beyond. In this work, the focus is on multi-RAT environments that enable the coexistence of B5G New Radio (NR) cellular infrastructures with complementary technologies such as Wireless Fidelity (WiFi) \cite{syllaMulticonnectivity, bartsiokas2025federated}. Most modern User Equipments (UEs) can simultaneously connect to both WiFi and 5G. In such dual-connectivity scenarios, UEs typically prioritize WiFi connections to minimize cellular data charges. However, this often leads to throughput degradation in areas with obstacles or poor WiFi coverage. To overcome this issue, handover mechanisms are employed to dynamically switch between WiFi and cellular networks according to a predefined selection policy \cite{yangSmartHO}.

Traditional handover mechanisms between different RATs in current 5G/NR systems are typically event-driven and rely on instantaneous measurements of the Received Signal Strength Indicator (RSSI) that compare serving and neighbor signal levels with thresholds and time-to-trigger (TTT) timers \cite{SATAPATHY2023adaptive}. Although conventional methods are simple and computationally efficient, they often result in suboptimal Quality of Service (QoS), especially in dynamic and heterogeneous environments. Conditional Handover (CHO), standardized in 3GPP Release 16 \cite{3GPP_TS_38300}, pre-configures one or more target cells together with execution conditions. The final handover is executed once those conditions are met, reducing command latency and improving reliability under fast fading and blockage. However, both conventional and CHO remain fundamentally reactive, since decisions are taken when thresholds are crossed, not based on forecasted link quality or under rapidly changing interference in multi-RAT settings. Recently, Machine Learning (ML) and Deep Learning (DL) have emerged as promising enablers for intelligent mobility management, aiming to overcome these limitations and enable proactive handover decisions \cite{giannopoulos2025ai, ALMEIDA2024ml}. Their ability to learn complex patterns makes them particularly suitable for predicting multi-source wireless signal quality. Furthermore, ML-based mobility policies can simultaneously exploit UE measurements, environmental conditions, and user-specific performance objectives, achieving a balance between connectivity and efficiency \cite{masri2021machine}.

In this context, this paper proposes a Predictive CHO (P-CHO) framework that supports ML-assisted dynamic RAT selection under mobility and interference conditions. The goal is to augment CHO with model-driven forecasts of near-future signal quality for candidate cells/RATs, turning execution conditions from static thresholds into time-aware predicates derived from predicted Signal-to-interference-plus-noise ratio (SINR) trajectories. By anticipating short-term degradations and opportunities (e.g., impending WiFi shadowing or inter-cellular interference), P-CHO enables earlier preparation of target links and smarter selection among multiple RATs. ML/DL models, especially sequence models such as Long Short-Term Memory (LSTM), are well-suited for this task because they (i) can capture temporal dependencies in radio dynamics, (ii) can provide accurate predictions of UE measurements, and, finally, (iii) generate actionable predictions for mobility control and traffic steering. The main contributions of this work are summarized as follows:

\begin{itemize}
    \item We formalize P-CHO for sixth-generation (6G) multi-RAT networks by integrating short-horizon SINR forecasts into CHO execution logic, enabling proactive vertical and horizontal handovers across Cellular and WiFi RATs.
    \item We design and train RAT-specific SINR forecasting models, including bidirectional LSTM (BiLSTM) for 6G Cellular BSs and a lightweight LSTM for WiFi APs, producing next-step and multi-step SINR estimates along stochastic user trajectories.
    \item We implement a hybrid Cellular/WiFi HetNet simulator that generates mobility and channel-realistic datasets (path loss, shadowing, small-scale fading, and interference) while the UE traverses predefined routes in dense deployments.
    \item We embed the SINR predictors into a RAT Steering Controller that combines multi-source UE measurement reports, infers pre-trained SINR prediction models, and converts SINR forecasts into steering execution conditions, leading to dynamic RAT-user associations.
    \item We benchmark the proposed LSTM-based signal predictors against other ML-based models and traditional autoregressive models, demonstrating consistent improvements under dynamic interference and mobility.
\end{itemize}

The rest of this manuscript is organized as follows: In Section II various previous research that has been done in this field is presented. Following is Section III, where the system model of a heterogeneous environment with the co-existence of 6G BSs and WiFi APs is formulated. In Section IV, the proposed ML/DL algorithms and P-CHO framework for intelligent signal prediction are presented, while in Section V the performance evaluation of the above is described. Finally, concluding remarks and future directions are provided in Section VI.

\section{Related Work}

Mobility management in 5G/6G HetNets has drawn significant attention, particularly at the intersection of handover control and data-driven intelligence. Most existing studies analyze handover decision‐making within 5G NR or across Long Term Evoluation (LTE)–5G, with increasing interest in predictive policies leveraging sequence models. In \cite{baz2024enhancing}, a deep residual matrix LSTM was proposed to predict future user locations and trigger handover when the distance to the serving Base Station (BS) exceeds a predefined threshold. The method achieved an increased handover success rate and reduced latency, outperforming Reinforcement Learning (RL) and the Adaptive Cell Selection Algorithm (ACSA) schemes. While compelling, the approach was confined to intra-5G scenarios and did not consider multi-RAT interactions. A complementary direction is taken in \cite{masri2021machine}, where an LSTM predicts the probability that each neighboring cell will have the highest Received Signal Strength Power (RSRP), formulating handover as a multi-class classification problem with a post-processing dynamic threshold to balance false positives/negatives. Evaluated in a simulated industrial 5G setup, the approach reduced radio link failures and mitigated ping-pong effects. However, the focus remained on 5G RSRP rather than cross-RAT signal dynamics.

Learning-based policy optimization has also been explored beyond pure prediction. The authors in \cite{song2023handover} studied dense HetNets with a deep Q-learning strategy that selects candidate BSs per time slot to optimize throughput, delay, and energy consumption. The RL method outperformed conventional A3 policies and decreased the energy usage (e.g., to 0.033,J/s in dynamic channels and 0.023,J/s in quasi-static channels), yet it did not explicitly forecast short-horizon signal evolution. In a related concept, the authors in \cite{lee2020prediction} extended the 3GPP-based CHO framework for millimeter-wave (mmWave) 5G with a deep neural model trained on geographical blockage patterns and RSRP sequences, producing next-cell probabilities to enable early preparation. The reported accuracy (85\%) and early-preparation success rate (98\%) showcased the value of preparing handovers ahead of execution events, but the study remained single-RAT and RSRP-centric.

Regarding WiFi-related works, the study in \cite{cervantes2022proactive} proposed a proactive cross-layer handover framework for Wireless Local Area Networks (WLANs) that extracts network-layer features (RSSI, retransmissions, packet delay) to train ML classifiers distinguishing handover from non-handover states. The method reduces handover latency by approximately 200 ms and packet loss by over 30\%. While effective, the study did not extend to heterogeneous cellular–WiFi coordination.

Works explicitly targeting multi-RAT are relatively sparse. One notable effort is \cite{yang2024smart}, which combines a Multi-acess Edge Computing (MEC)-based throughput estimator with Deep Reinforcement Learning (DRL) to learn discrepancies between estimated and actual throughput, thereby guiding the selection between 5G and WiFi transmitters and using Quick UDP Internet Connections (QUIC) scheme for migration. Nevertheless, the DRL agent requires periodical re-training once the network dynamics significantly change, leading to time-consuming model adjustment periods.

Prior studies either (i) operate within a single RAT with RSRP-based targets \cite{baz2024enhancing,masri2021machine,lee2020prediction}, (ii) optimize policies without explicit short-horizon signal forecasting \cite{song2023handover}, or (iii) treat multi-RAT selection primarily as throughput-driven migration \cite{yang2024smart} or WiFi-only handover \cite{cervantes2022proactive}. To fill these gaps and extend the existing work to hybrid cellular/WiFi networks, this work targets predictive multi-RAT mobility by forecasting RAT-specific SINR along user trajectories and embedding these forecasts into a P-CHO logic.

\section{Multi-RAT System Model}

This section presents the proposed system model for proactive handover in a hybrid cellular/WiFi multi-RAT HetNet. Section~\ref{AA} describes the considered network topology, system entities, channel models and user mobility patterns. Section~\ref{BB} formulates the key problem elements for signal quality prediction-based handover. The core objective is to forecast the experienced time-varying SINR metrics for each RAT transmitter as users move across mobility trajectories and, then, determine the optimal target RAT.

\subsection{System Overview}\label{AA}

\begin{figure}
\centerline{\includegraphics[width=3.5in]{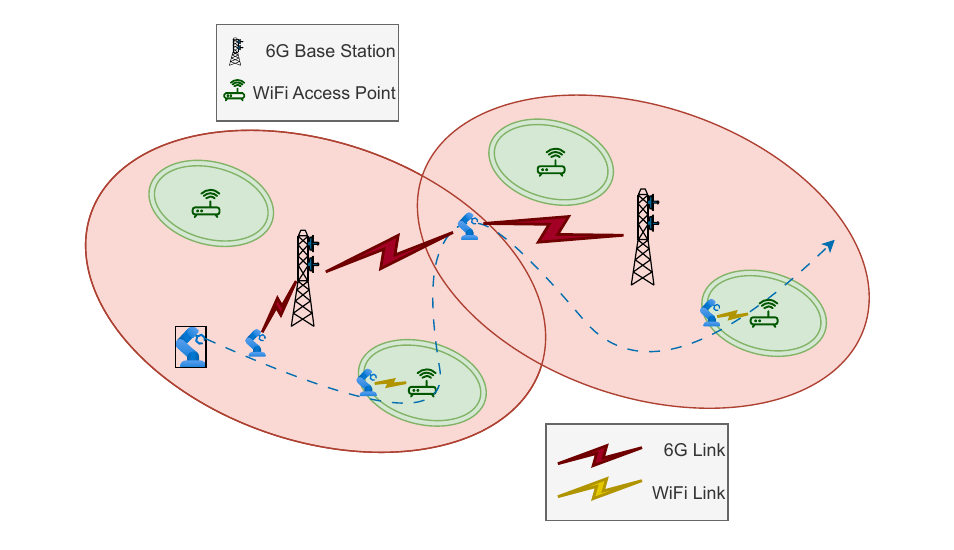}}
\caption{Multi-RAT HetNet with co-existing cellular/WiFi coverage. Dashed blue curve reflects a UE random-walk path.
\label{figure_1}}
\end{figure}

We consider a heterogeneous multi-RAT environment representative of B5G/6G deployments. The network comprises wide-area cellular BSs and overlapping short-coverage WiFi Access Points (APs), with mobile UEs traversing predefined paths within the multi-RAT coverage area. Fig.~\ref{figure_1} illustrates an example layout in which user trajectories traverse overlapping coverage regions of multiple RATs. We define the following sets: (i) Set of cellular BSs $S_{BS} = \{b_1, b_2, \dots, b_{N_{BS}}\}$ (includes $N_{BS}$ BSs), (ii) Set of APs $S_{AP} = \{a_1, a_2, \dots, a_{N_{AP}}\}$ (includes $N_{AP}$ APs each one located within a specific BS's coverage), (iii) Set of UEs $S_{UE} = \{u_1, u_2, \dots, u_{N_{UE}}\}$ (includes $N_{UE}$ UEs moving with a certain velocity), and (iv) Set of UE paths $S_T = \{t_1, t_2, \dots, t_{N_T}\}$ (includes $N_{T}$ UE trajectories created as polynomial interpolations, representing random walks across the whole multi-RAT area).

Two link types are considered, including (i) $L_{b,u}$ for cellular BS-UE communication and (ii) $L_{a,u}$ for AP–UE communication. Each link employs a RAT-specific channel model. For cellular links $L_{b,u}$, the received power accounts for large-scale path loss, small-scale fading, and co-channel interference from neighboring BSs. A generic distance-based attenuation is defined as the large-scale path loss component
:

\begin{equation}
G(d) = K \cdot d^{-\alpha}
\label{eq:pathloss_cell}
\end{equation}

\noindent where $d$ is the BS–UE distance, $\alpha$ is the path-loss exponent, and $K$ a scaling constant \cite{giannopoulosspatiotemporal}. Small-scale fading is modeled as Rician to capture both Line-of-Sight (LoS) and non-LoS (NLoS) components. For a certain UE, let $h$ denote the complex fading coefficient experienced over the serving link, and $h_i$ denote the same for the interfering link $i$. The instantaneous cellular SINR is then expressed as:

\begin{equation}
\mathrm{SINR}_{6G} = \frac{P_t \cdot |h|^2} {\sum_{i \in I} P_{t,i} \cdot |h_i|^2 + N_0 \cdot B} 
\label{eq:sinr_cell}
\end{equation}

\noindent where \(I\) is the set of interfering BSs, $P_t$ ($P_{t,i}$) is the serving (cellular interferer's $i$) transmit power, \(N_0\) is the spectral density of the noise power and \(B\) is the link bandwidth. The channel model of WiFi links \(L_{a,u}\) follows an IEEE 802.11 channel model \cite{capulli2006path} that captures short-range indoor hotspot communications reflecting multipath and wall penetration losses:

\begin{equation}
PL(d) = PL(d_0) + 10 \cdot \gamma \cdot log_{10}{\frac{d}{d_0}} + \sum_{j}L_j
\label{eq:pl_wifi}
\end{equation}

\noindent where \(PL(d_0)\) stands for the reference path loss at distance $d_0=1$ m, $\gamma$ is the indoor path loss exponent, and $L_j$ represents additional loss due to obstacle $j$ (e.g. wall) \cite{capulli2006path}. Small-scale fading and interference are omitted in this case, resulting in an SNR formula:

\begin{equation}
\mathrm{SNR}_{WiFi} = \frac{P_t \cdot |h|^2} {N_0 \cdot B}
\label{eq:snr_wifi}
\end{equation}

UEs move along predefined two-dimensional (2D) trajectories $t \in S_T$ with constant or piecewise-constant speed. Trajectories are selected to span diverse directions, turning points, and RAT overlaps, thereby inducing varied sequences of serving RATs. The sampling interval determines the number of trajectory points between start and end locations for a given UE speed.

\subsection{Handover Decision Problem Elements}\label{BB}

Given a trajectory $t$ and the network state at discrete time $k$, let $\mathbf{x}_k$ denote the feature vector (e.g., past per-RAT signal measurements, position/velocity values, RAT identifiers, or environmental indicators). The goal is to learn the following signal quality predictors:

\begin{equation}\label{eq:pred}
\hat{s}^{(r)}_{k+\tau} = f^{(r)}(\mathbf{x}_k), \quad r\in\{S_{BS} \cup S_{AP}\},\;\; \tau\in\{1,\dots,H\},
\end{equation}

\noindent where $\hat{s}^{(r)}_{k+\tau}$ is the predicted signal quality indicator for time instance $k+\tau$ and RAT $r$, $f^{(r)}$ is the learned function for RAT $r$ that maps historical to future signal quality metrics, and $\tau$ is the forecasting horizon (e.g., 3 steps ahead). Predictions $\hat{s}^{(r)}_{k+\tau}$ estimate short-horizon signal quality for each RAT using a historical feature vector $\mathbf{x}_k$ of $\mathrm{SINR}_{\mathrm{6G}}$ values (for cellular BSs) and $\mathrm{SNR}_{\mathrm{WiFi}}$ values (for WiFi APs). At each step $k$, the RAT steering controller selects the target serving RAT as:

\begin{equation}
r_k^\star=\arg\max_{r} \hat{s}^{(r)}_{k+\tau}
\end{equation}

\noindent where the decision combines the predicted signal quality from all transmitters and selects the best-SINR provider. The final decision can be conditioned by rule-based policies (e.g., trigger a handover when a certain BS switch gain is met) using multi-step predictors $\{\hat{s}^{(r)}_{k+\tau}\}_{\tau=1}^{H}$.

\section{ML-assisted Proactive Handover}

This section details the proposed scheme for intelligent proactive handover in a heterogeneous multi-RAT network using ML/DL models. We first describe the dataset construction pipeline, covering the simulation process, sequential data generation, and data organization, and then outline the ML/DL architecture employed for prediction. At the end of this section the proposed workflow is described in detail.

\subsection{Network Simulator and Dataset Construction}

Obtaining recent, real, and authenticated communication traces is challenging due to privacy policies and personal data regulations. Yet, ML/DL models require structured data for training and evaluation. To this end, we developed a Python-based simulator for the 5G/WiFi HetNet that represents the targeted communication environment and exports model-ready sequential datasets.

As outlined in Section~\ref{AA}, we develop a simulator that establishes communication between all available RAT endpoints and the UE while the latter moves along a predefined trajectory $t\in S_T$. For each trajectory, and at each discrete time index $k$, the simulator evaluates per-candidate serving transmitter metrics for all $r \in R$, where $R \triangleq S_{\!BS}\cup S_{\!AP}$.

Specifically, for every time instance of the user mobility path and BS/AP candidate, the simulator computes the RSSI, the SINR, and the achieved throughput (denoted as $\mathrm{TP}$). The resulting raw time series are temporally aligned across RATs and stored as tuples in the form of $\mathrm{M}_{r}^{t}[k] = (\mathrm{ID}_{t}, \mathrm{RSSI}_r[k], \mathrm{SINR}_r[k], \mathrm{TP}_r[k])$, where $\mathrm{ID}_{t}$ is the identifier of trajectory $t$. All the other metrics refer to the $RSSI$, $SINR$ and $TP$ values measured for transmitter $r \in R$ at time instance $k$ following trajectory $t$. This temporal alignment of the multi-source timeseries ensures that, for each time instance $k$, measurements across RATs refer to the same UE position/sample.

To create input-output combinations for subsequent model training, the raw sequences are then segmented into overlapping windows of fixed length $W$. For node $r$ and time index $k$, the input feature vector is:

\begin{equation}
\mathbf{x}_{r}[k]=\big[\;\mathrm{M}_{r}^{t}[k-W+1],\,\ldots,\,\mathrm{M}_{r}^{t}[k]\;\big], \forall t \in S_T
\label{eq:window_def}
\end{equation}

\noindent where $\mathrm{M}_{r}^{t}[k]$ is the metric report tuple for node $r$ when UE follows trajectory $t$, and is used to make a prediction at time $k$. This means that, to produce a prediction at time $k$, the model receives a history window of the $W$ previous metric report tuples. At time $k$, the one-step-ahead target $y_{r}[k]$ (i.e., the model's prediction) is the (unknown) upcoming metric tuple of the next time instance, i.e., $y_{r}[k]=\mathrm{M}_{r}^{t}[k+1]$.

The dataset is organized in a tabular form with columns $\mathrm{ID}_{t}$, $r$, $\mathbf{x}_{r}[k]$ (model inputs), and $y_{r}[k]$ (model desired output), and exported into two separate (\texttt{.csv}) files for model training purposes. We train two models, including one for cellular BSs samples and one for APs samples. These datasets serve as inputs to the LSTM-based predictors described in Section \ref{sec:models}. Standard sequence-model preprocessing steps (e.g., normalization, windowing, and train/test partitioning) are applied to ensure consistent scaling and robust evaluation.

\subsection{ML Model Architecture for Signal Quality Prediction}\label{sec:models}

We employ LSTM networks to forecast short-horizon signal quality for candidate BSs and APs. LSTMs, a class of recurrent neural networks, are designed to capture long- and short-range temporal dependencies in sequential data and are therefore well suited to regress wireless signal evolution under mobility and time-varying interference.

\begin{figure}
\centerline{\includegraphics[width=3.5in]{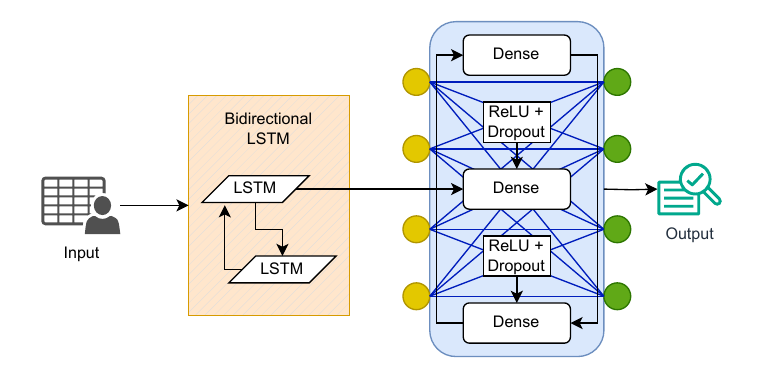}}
\caption{LSTM structure for timeseries signal quality prediction.\label{figure_2}}
\end{figure}

\paragraph*{Cellular BS predictor (BiLSTM)}
Cellular BS-UE links exhibit richer temporal structure due to path-loss variation in urban environments, blockage, and co-channel interference with adjacent BSs. To capture these dynamics, we use a deep BiLSTM. As shown in Fig.~\ref{figure_2}, the first block processes the input window with two stacked LSTM layers in both forward and backward directions, enabling the model to exploit information across the entire input sequence. The recurrent output is then fed to a stack of fully-connected dense layers interleaved with ReLU activations and dropout to avoid overfitting. The output is a linear neuron that outputs a scalar next-step prediction $\hat{y}_{r}[k]$ for the selected metric.

\paragraph*{WiFi AP predictor (lightweight LSTM)}
$L_{a,u}$ links typically exhibit shorter-range propagation with simpler temporal variability due to absence of interference (no AP-BS interference). We therefore adopt a lightweight design with a single LSTM layer followed by one Dense layer that maps the last hidden state to the scalar output $\hat{y}_{r}[k]$. This design reduces computational and memory footprint while retaining sufficient accuracy for short-horizon prediction.

\paragraph*{Training and objective}
Models are trained offline using mean squared error (MSE)-based loss function. During online operation, a RAT Steering Controller queries the appropriate predictor (BiLSTM for BSs, lightweight LSTM for APs) to produce per-RAT short-horizon forecasts that feed the predictive P-CHO logic described below.

\subsection{Proposed P-CHO Sequence Diagram}

\begin{figure}
\centerline{\includegraphics[width=3.5in]{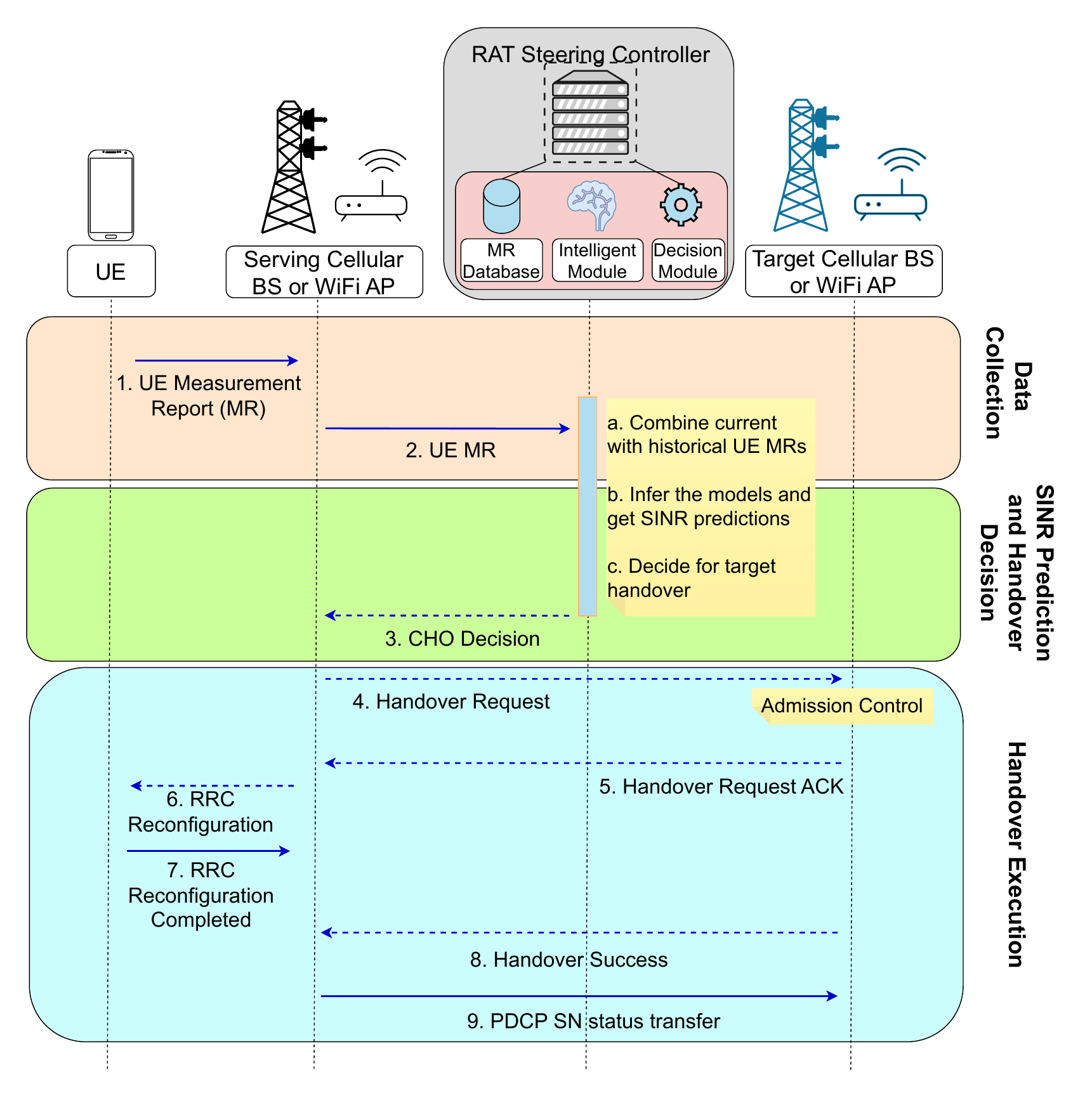}}
\caption{Proposed three-stage P-CHO sequence callflows. \label{figure_3}}
\end{figure}

Recent 3GPP releases introduce CHO, where preparation and execution are decoupled and execution occurs only when a condition is satisfied \cite{3GPP_TS_38300}. We extend CHO to a P-CHO by deriving the condition from short-horizon, model-based SINR forecasts, thereby enabling proactive mobility, as shown in Fig.~\ref{figure_3}.

\paragraph*{Prediction-conditioned trigger}
Let $\hat{s}_{\mathrm{tar}}[k]$ and $\hat{s}_{\mathrm{cur}}[k]$ denote the one-step-ahead predicted SINR (or SNR for WiFi) for the candidate target RAT and the current serving RAT, respectively. The execution condition is:

\begin{equation}
\Delta \mathrm{QoS}[k]
\;\triangleq\;
\hat{s}_{\mathrm{tar}}[k]-\hat{s}_{\mathrm{cur}}[k]
\;\ge\;
\delta_{\mathrm{QoS}}
\label{eq:pcho_trigger}
\end{equation}

\noindent where $\delta_{\mathrm{QoS}}$ is a configurable threshold that reflects whether the handover is soft (low $\delta_{\mathrm{QoS}}$ is sensitive to frequent switching) or hard (high $\delta_{\mathrm{QoS}}$ avoids ping-pong). By combining ML-based SINR prediction with handover sensitivity thresholds, the system adopts a proactive and predictive behavior, allowing early preparation of handover events while regulating handover sensitivity \cite{park2024proactive}. 

\paragraph*{End-to-end workflow (mapped to Fig.~\ref{figure_3})} The P-CHO procedure comprises three stages:

\textbf{(1) Data Collection:} The UE transmits a Measurement Report (MR) via the serving BS/AP (RSRP, SINR/SNR, throughput). The RAT Steering Controller stores the MR in an MR timeseries database and associates it with the UE’s historical MR sequence.

\textbf{(2) SINR Prediction \& Handover Decision:} Then, batch feature assembly operation is performed by the RAT Steering Controller, in which current MR features are combined with the UE’s recent window to form inputs $\mathbf{x}_{r}[k]$ for each candidate RAT $r \in R$. A parallel inference of SINR prediction models is then performed in the Intelligent Module. Specifically, it infers the pre-trained predictors (BiLSTM for BSs, lightweight LSTM for APs) in parallel to obtain $\{\hat{s}^{(r)}[k]\}$ for $r\in R$. Optionally, model inference can be done on Graphical Processing Unit (GPU) to further speed up the process. Afterwards, the Decision Module evaluates the condition in \eqref{eq:pcho_trigger} for each candidate. If the condition holds for some $r^\star \in R$ (target BS/AP), it issues a P-CHO decision toward $r^\star$. Otherwise, the UE remains on the serving RAT.

\textbf{(3) Handover Execution (conditional):} Following a 3GPP-compliant procedure, based on the decision, the serving node sends a Handover Request to the target BS/AP $r^\star$. Upon request receipt, the target node performs admission control (resource availability/QoS checks) and responds with Handover Request ACK on success. Then, the serving node issues Radio Resource Control (RRC) Reconfiguration to the UE with target parameters. Upon completion, the UE synchronizes to the target. Finally, the status notification is transferred (e.g., PDCP Sequence Numbers) to finalize context continuity and confirm handover success.

\section{Performance Evaluation}

We consider a heterogeneous multi-RAT Cellular/WiFi network with mobile users served by BSs or APs. As in the topology of Fig.~\ref{figure_1}, the cellular tier comprises two B5G NR BSs operating at 3.5 GHz with partially overlapping cells. Inter-cell interference is present near the cell edges. Within each BS coverage, two IEEE 802.11 APs operate at 2.4 GHz on orthogonal channels (no co-channel interference across APs, no BS-AP interference due to different frequency bands). UEs traverse the topology along predefined trajectories at a constant speed of $3$ m/s. As described in Sections~\ref{AA} and~\ref{BB}, the Python-based link/system simulator generates time-synchronized sequences of per-RAT measurements (RSSI, SINR/SNR, throughput) for downlink traffic. 

The problem of SINR prediction is examined as a supervised timeseries regression problem. We consider short-horizon signal prediction as a sequence-to-one regression problem (one-step-ahead unless otherwise stated). For each candidate RAT endpoint $r$, a sliding window of length $W$ forms the input, and the next-sample value of the SINR forms the target (see \eqref{eq:window_def}). A training/test set split of 80\%/20\%  has been used in the training phase of all approaches, while the models were tuned via grid search on different hyperparameters to ensure optimal performance.

\subsection{Prediction Error vs Look-back and Look-ahead Windows}

\begin{figure}[t!]
\centering
\begin{minipage}[t]{0.25\textwidth}
\centering
\includegraphics[width=\textwidth,height=3cm]{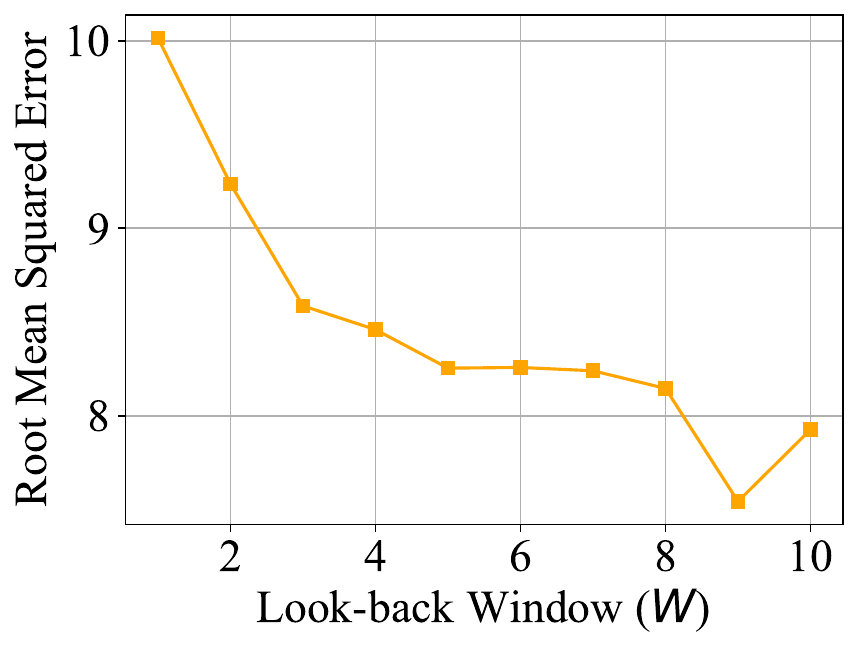} 
\subcaption{}
\label{fig:subim1}
\end{minipage}%
\begin{minipage}[t]{0.25\textwidth}
\centering
\includegraphics[width=\textwidth,height=3cm]{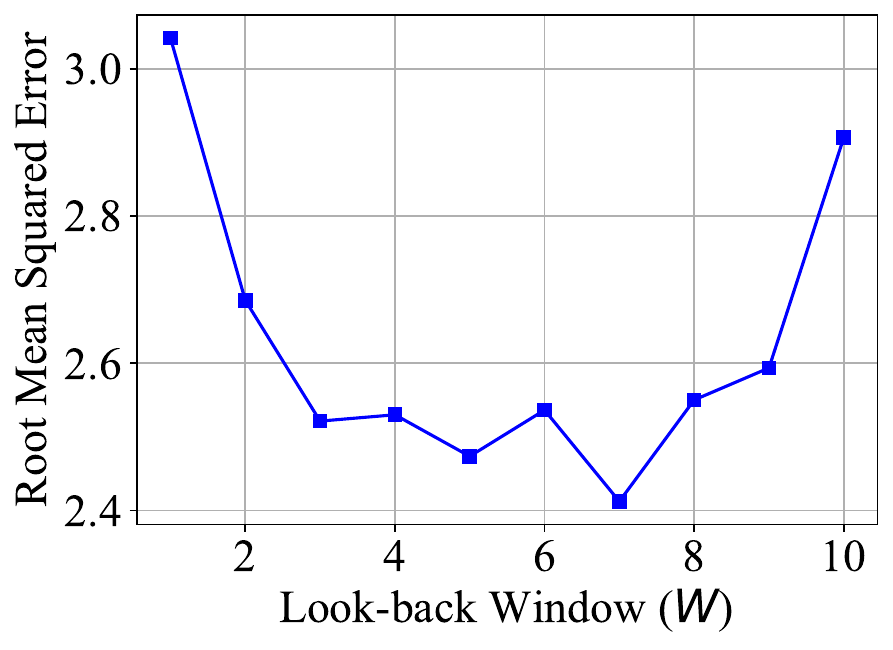}
\subcaption{}
\label{fig:subim2}
\end{minipage}
\caption{Prediction RMSE versus Look-back Window size $W$ for (a) the Cellular BSs model and (b) WiFi APs model, considering 35 available UE paths.}
\label{fig:image3}
\end{figure}

Performance evaluation was conducted through a series of experiments under varying conditions and objectives. First, we examined the impact of the lookback window size $W$ on SINR prediction error (between actual and predicted SINR values of the testing set). As shown in Fig.~\ref{fig:image3}, increasing the window generally improves performance. For BS models, the trend is consistent, with the best Root Mean Squared Error (RMSE) achieved at a window of $W_{BS}=9$ past time steps. For AP models, the curve is U-shaped, indicating that long windows lead to overfitting. Thus, the optimal window was set at $W_{AP}=7$. 
 
\begin{figure}[t!]
\centering
\begin{minipage}[t]{0.25\textwidth}
\centering
\includegraphics[width=\textwidth,height=3cm]{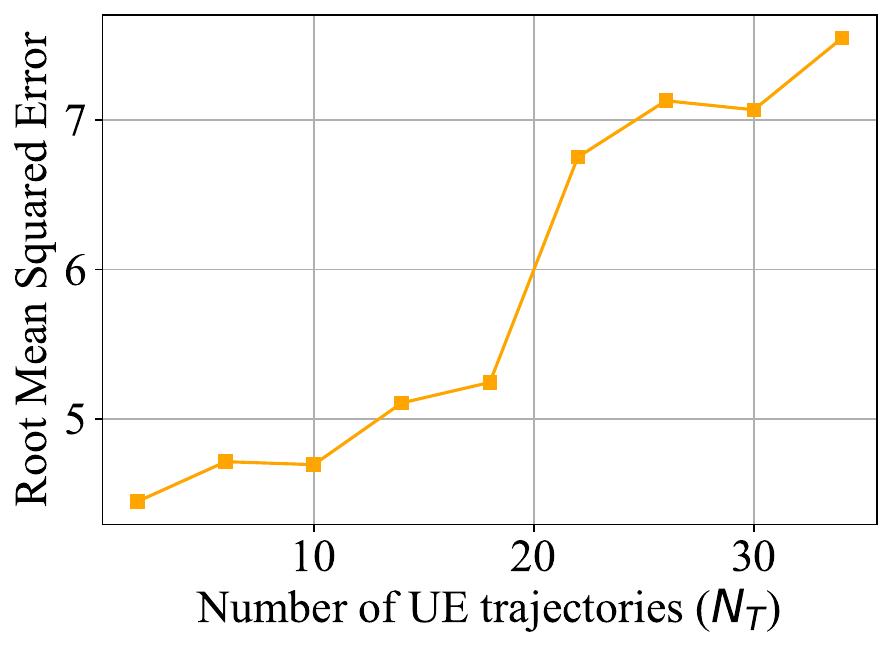} 
\subcaption{}
\label{fig:subim3}
\end{minipage}%
\begin{minipage}[t]{0.25\textwidth}
\centering
\includegraphics[width=\textwidth,height=3cm]{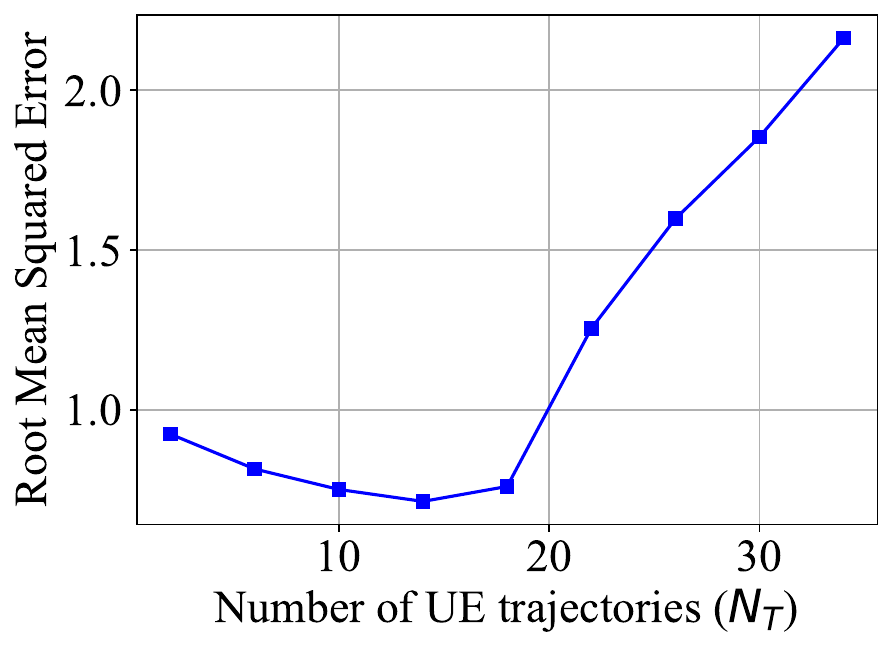}
\subcaption{}
\label{fig:subim4}
\end{minipage}
\caption{Prediction RMSE versus the number of UE paths $N_T$ for (a) the Cellular BSs model and (b) WiFi APs model.}
\label{fig:image4}
\end{figure}

Next, we investigated the effect of the number of trajectories available for each UE. In general, adding more available random-walk paths increases the system stochasticity and makes the SINR prediction task more difficult. This is because, with a fixed topology, adding more trajectories increases route overlap among UEs, which tends to confuse the models and degrades their ability to discriminate among paths.  As illustrated in Fig.~\ref{fig:image4}, enlarging the training set with additional trajectories does not monotonically improve generalization under a fixed topology. For BS models, RMSE remains relatively stable up to $N_T = 15$ trajectories and then increases sharply beyond $N_T = 20$. For AP models, error decreases until $N_T = 15$ trajectories (benefiting from added variability) but rises thereafter, suggesting SINR distribution shift. To ensure a balance between adequate model performance and system stochasticity, we set the number of trajectories at $N_T=20$ for the rest of the simulations. Note that, to increase model performance under large values of $T$, more dense (i.e., with more hidden layers) LSTM models can be retrained for better accuracy outcomes.

\begin{figure}[t!]
\centering
\begin{minipage}[t]{0.25\textwidth}
\centering
\includegraphics[width=\textwidth,height=3cm]{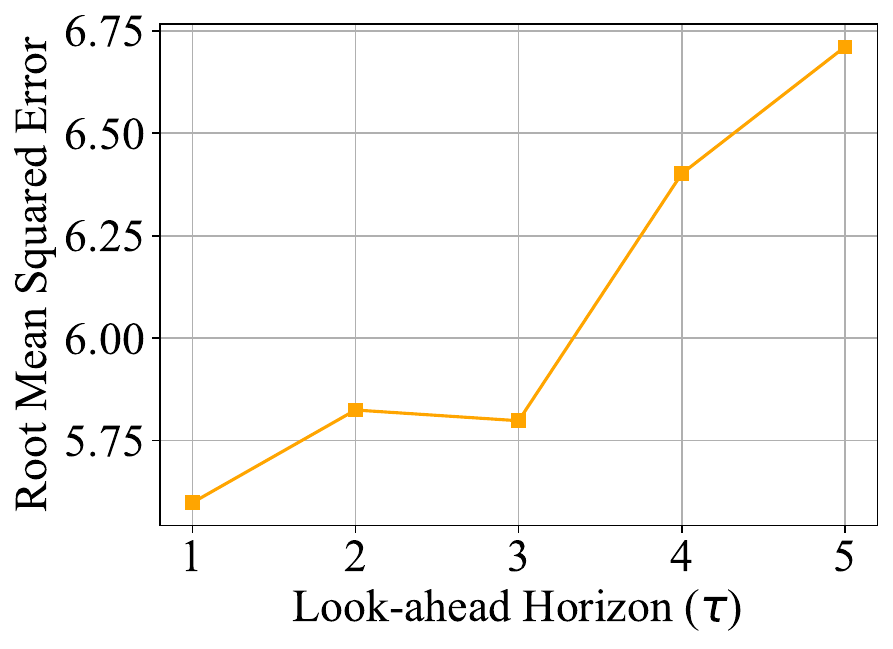} 
\subcaption{}
\label{fig:subim5}
\end{minipage}%
\begin{minipage}[t]{0.25\textwidth}
\centering
\includegraphics[width=\textwidth,height=3cm]{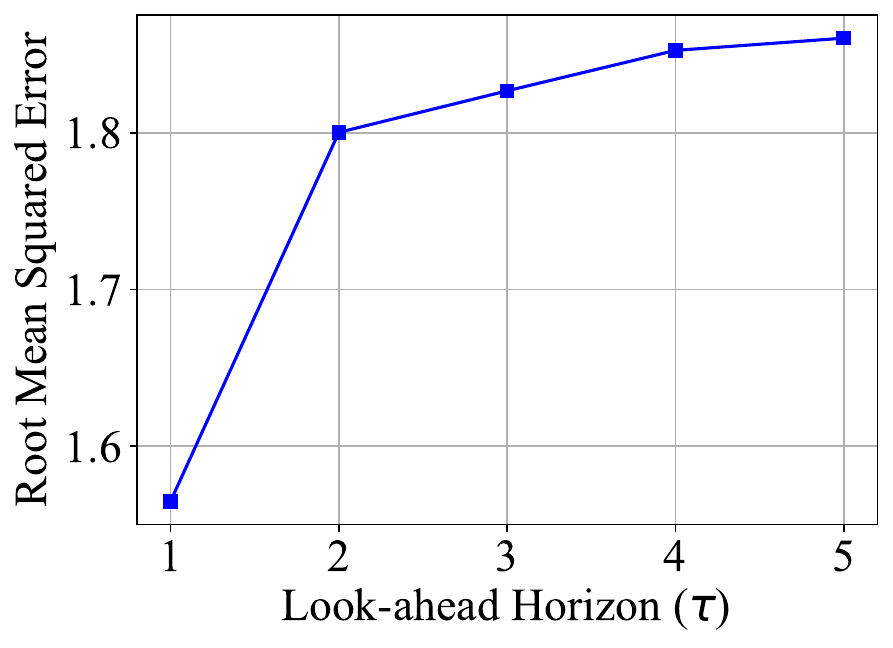}
\subcaption{}
\label{fig:subim6}
\end{minipage}
\caption{Prediction RMSE versus Look-ahead Horizon $\tau$ in multi-step prediction for (a) the Cellular BSs model and (b) WiFi APs model.}
\label{fig:image5}
\end{figure}

We further evaluated the predictive capacity of the models by extending the forecast horizon $\tau$ to multiple time steps. Two approaches were considered: \emph{(i) Direct multi-step:} the output layer is modified to emit $\tau=\{1,2,\dots,5\}$ steps jointly (see \eqref{eq:pred} for $H=5$), and the dataset is extended with targets $y_{r}[k]=\mathrm{M}_r^t[k+\tau]$. This means  that the prediction $y_{r}[k]$ made at time $k$ refers to the estimated SINR value at time $k+\tau$, with the models directly trained using the SINR desired values for $k+\tau$.  
\emph{(ii) Recursive (iterative) prediction:} the original single-step models are retained and used recursively, feeding each predicted value back into the input window for the next step. As expected, considering the direct multi-step scheme, Fig.~\ref{fig:image5} confirms that increasing the lookahead horizon $\tau$ generally degrades accuracy (for both BS and AP predictors) due to higher uncertainty. However, the error increment is moderate (particularly for the AP model), highlighting a practical accuracy–latency trade-off for decision making. This means that, if the look-ahead horizon is set to $\tau>1$ for early decisions, we may expect a lower prediction accuracy, but the handover operation will be fast without service downtime.

\begin{figure}[t!]
\centering
\begin{minipage}[t]{0.25\textwidth}
\centering
\includegraphics[width=\textwidth,height=3cm]{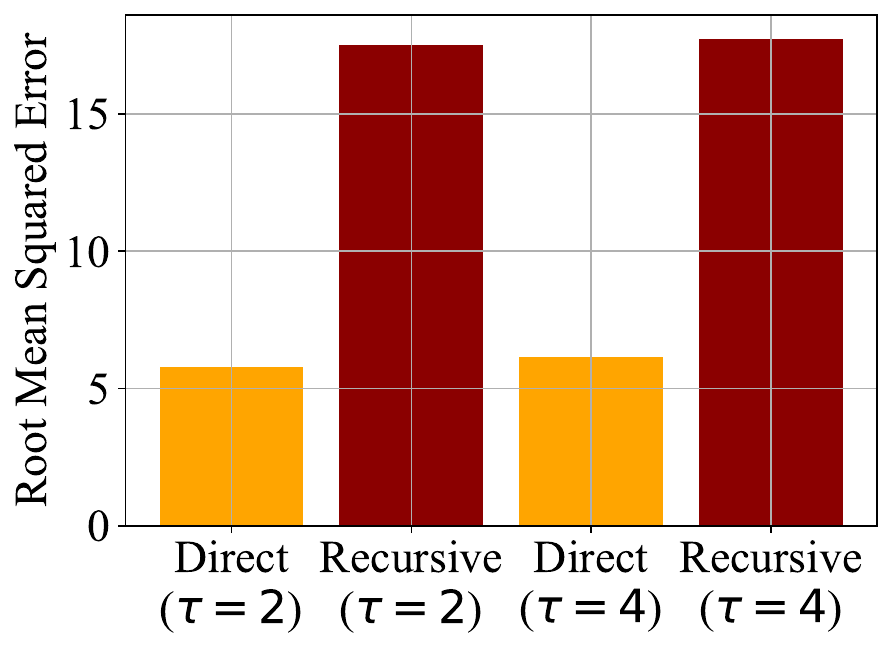} 
\subcaption{}
\label{fig:subim7}
\end{minipage}%
\begin{minipage}[t]{0.25\textwidth}
\centering
\includegraphics[width=\textwidth,height=3cm]{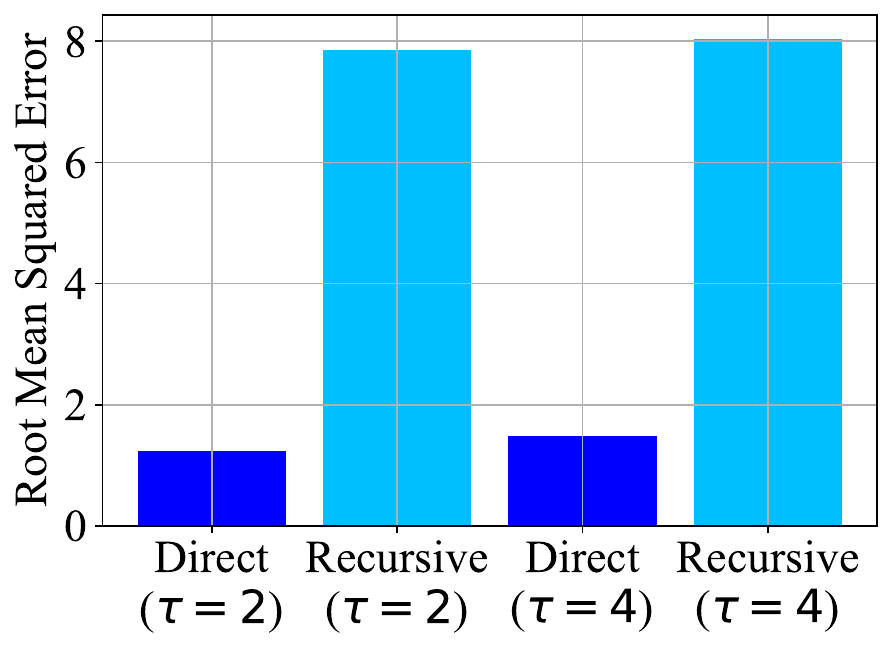}
\subcaption{}
\label{fig:subim8}
\end{minipage}
\caption{Prediction RMSE of Direct and Recursive multi-step methods using look-ahead horizons $\tau=2$ and $\tau=4$, separately for (a) Cellular BSs model and (b) WiFi APs model.}
\label{fig:image6}
\end{figure}

In Fig.~\ref{fig:image6}, we contrast \emph{direct multi-step} versus \emph{recursive} forecasting at lookahead $\tau=2$ and $\tau=4$ time steps. Direct multi-step yields consistently lower RMSE, with only a mild increase when moving from $\tau=2$ to $\tau=4$ for both BS and AP models. In contrast, the recursive scheme incurs substantial RMSE by roughly $3\times$ for BSs and $4$–$5\times$ for APs, relative to the direct scheme at the same horizon. Operationally, direct multi-step training requires storing multiple horizon-specific models at the RAT Steering Controller (per RAT), while the recursive scheme reuses only one single-step model, reducing storage at the expense of accuracy.

\subsection{Comparison against Baseline Signal Quality Predictors}

\begin{figure}[t!]
\centering
\begin{minipage}[t]{0.25\textwidth}
\centering
\includegraphics[width=\textwidth,height=3cm]{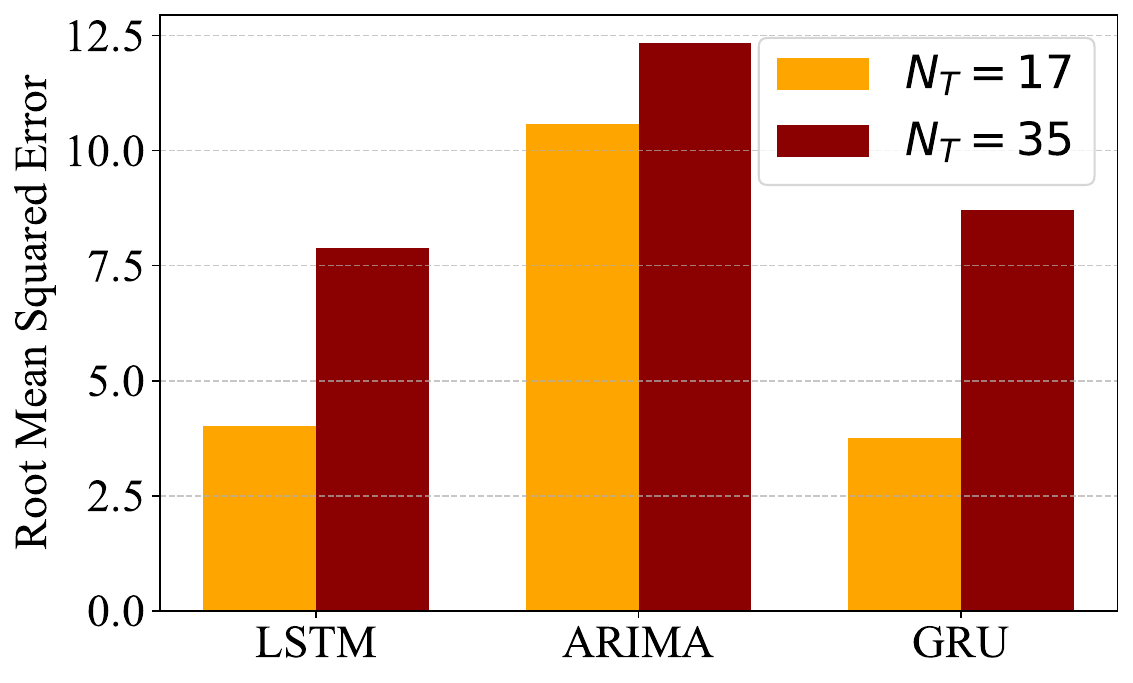} 
\subcaption{}
\label{fig:subim9}
\end{minipage}%
\begin{minipage}[t]{0.25\textwidth}
\centering
\includegraphics[width=\textwidth,height=3cm]{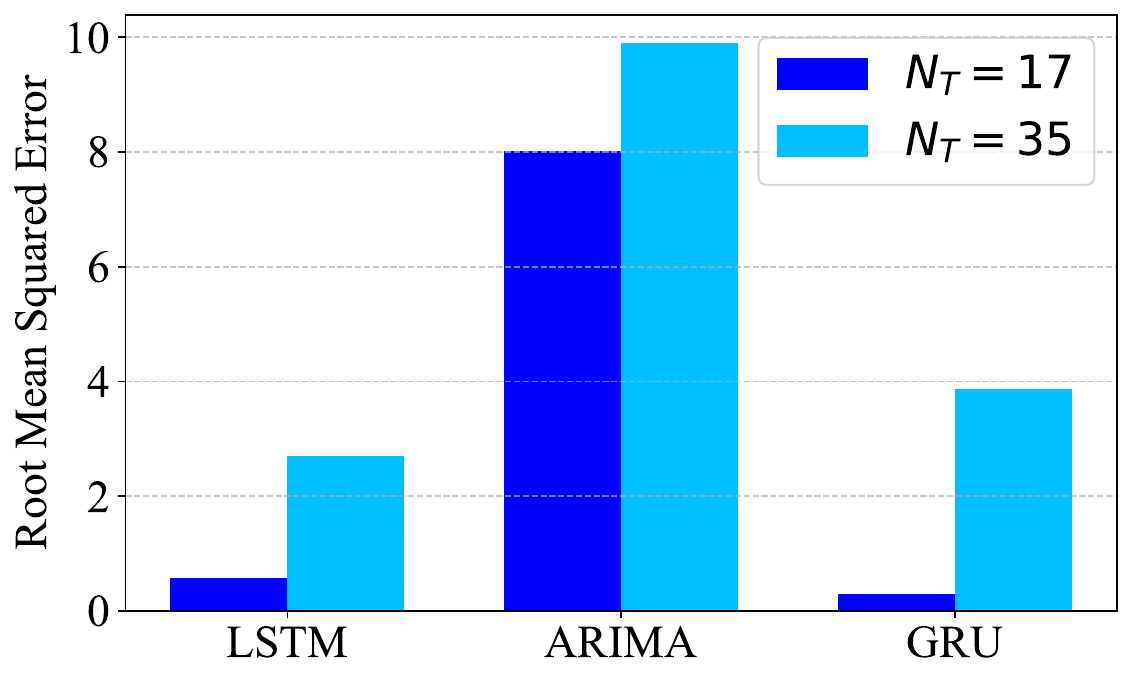}
\subcaption{}
\label{fig:subim10}
\end{minipage}
\caption{Prediction RMSE of different SINR predictors using different number of trajectories $N_T$ for (a) the Cellurar BSs models and (b) WiFi APs models.}
\label{fig:image7}
\end{figure}

Here we benchmark the proposed LSTM predictors against two classical baselines used for timeseries forecasting, including the \emph{Autoregressive Integrated Moving Average (ARIMA)} \cite{arima} and \emph{Extreme Gradient Boosting (XGBoost)} \cite{xgboost} models, under two $N_T$ configurations (number of UE trajectories $N_T=17$ and $N_T=35$). As shown in Fig.~\ref{fig:image7}, for each model, the RMSE is higher for $N_T=35$ due to the increased randomness in the mobility patterns. ARIMA exhibits the highest error than ML predictors across both RAT types and dataset scales, reflecting its linear/stationary assumptions and limited ability to model non-linear interference and mobility effects in multi-RAT links. Moreover, XGBoost achieves slightly lower prediction error than LSTM only for low number of UE paths ($N_T=17$), especially for AP prediction where short-range dynamics are simpler. However, its RMSE increases noticeably (relative to LSTM) when $N_T=35$, indicating sensitivity to trajectory overlap and mobility randomness. In contrast, LSTM predictors (i) maintain a comparable error for low $N_T$, relative to that of XGBoost, and (ii) clearly exhibit the lowest RMSE as $N_T$ increases, supporting its beneficial use when mobility patterns become more stochastic. Overall, the comparative results of Fig.~\ref{fig:image7} suggest that the proposed P-CHO pipeline may use LSTM or XGBoost predictors when mobility patterns are deterministic, whereas LSTM models should be used for more stochastic UE mobility patterns.

\subsection{Soft P-CHO versus Hysteresis-enabled P-CHO}

We evaluate two alternative realizations of the proposed P-CHO framework, including a \emph{Soft P-CHO} and a \emph{Hysteresis-enabled P-CHO} variant:
\begin{itemize}
  \item \textbf{Soft P-CHO:} The RAT Steering Controller triggers a handover when the next-step predicted gain (see \eqref{eq:pcho_trigger}) over the serving RAT exceeds the QoS threshold at the current step. This means that the UE will switch from the current to the target serving transmitter if the following Soft P-CHO condition is met:
  
\begin{equation}
    \Delta \mathrm{QoS}[k{+}1] \triangleq \hat{s}_{\mathrm{tar}}[k{+}1]-\hat{s}_{\mathrm{cur}}[k{+}1] \;\ge\; \delta_{\mathrm{QoS}}
    \label{eq:soft}
\end{equation}

  \noindent where $\hat{s}_{\mathrm{tar}}[k{+}1]$ and $\hat{s}_{\mathrm{cur}}[k{+}1]$ refer to the predicted SINR values from the current and the target RAT (respectively) at time instance $k+1$.
  \item \textbf{Hysteresis-enabled P-CHO.} The same condition must hold for $N$ consecutive steps (here $N \in \{2,3\}$) before executing the handover. This gives the following Hysteresis-enabled P-CHO condition:
  
\begin{equation}
  \Delta \mathrm{QoS}[h] \ge \delta_{\mathrm{QoS}}\;,\qquad \forall h \in [k{+}1,\,k{+}N]
    \label{eq:hard}
\end{equation}
  
  \noindent where $h$ reflects a hysteresis variable (in steps) denoting how many steps the target transmitter must comply with \eqref{eq:soft}. This means that hysteresis mechanism can be interpreted as a short dwell-time guard to mitigate ping-pong effects.
\end{itemize}

\begin{figure}[t!]
\centerline{\includegraphics[width=0.9\columnwidth]{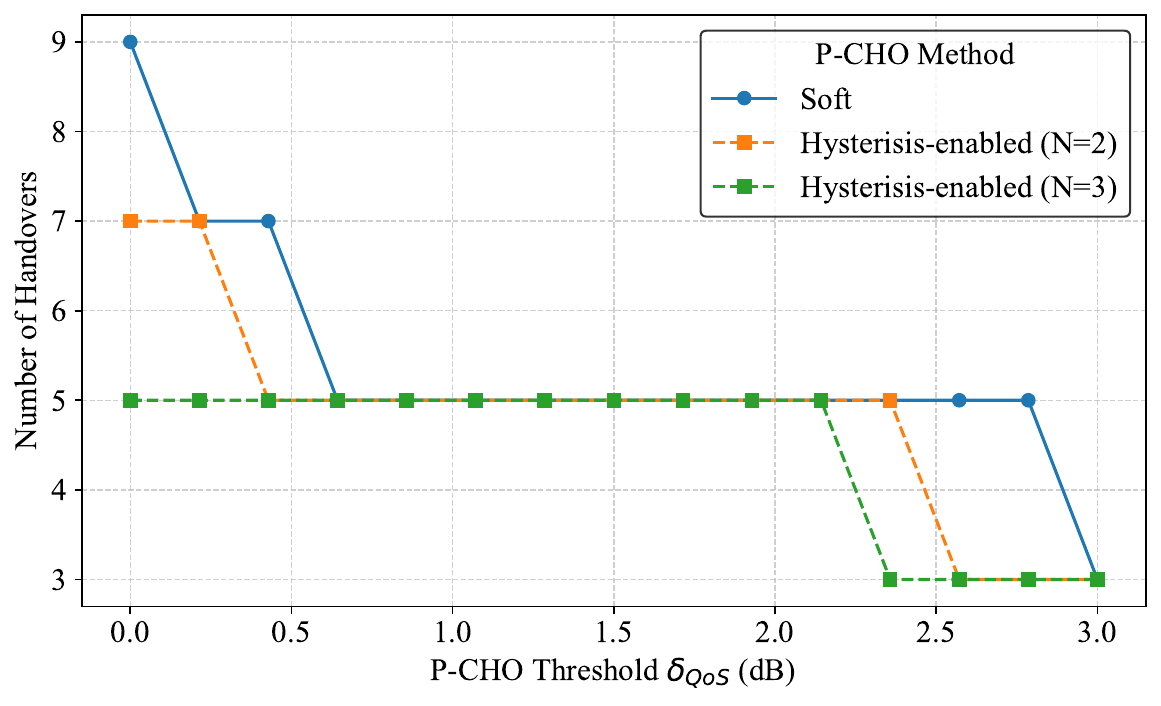}}
\caption{Number of Handover events as function of P-CHO Threshold $\delta_{\mathrm{QoS}}$ for Soft and Hysteresis-enabled methods.
\label{figure_8}}
\end{figure}

\begin{figure*}[t!]
\centering
\begin{minipage}[t]{0.33\textwidth}
\centering
\includegraphics[width=\textwidth,height=4.7cm]{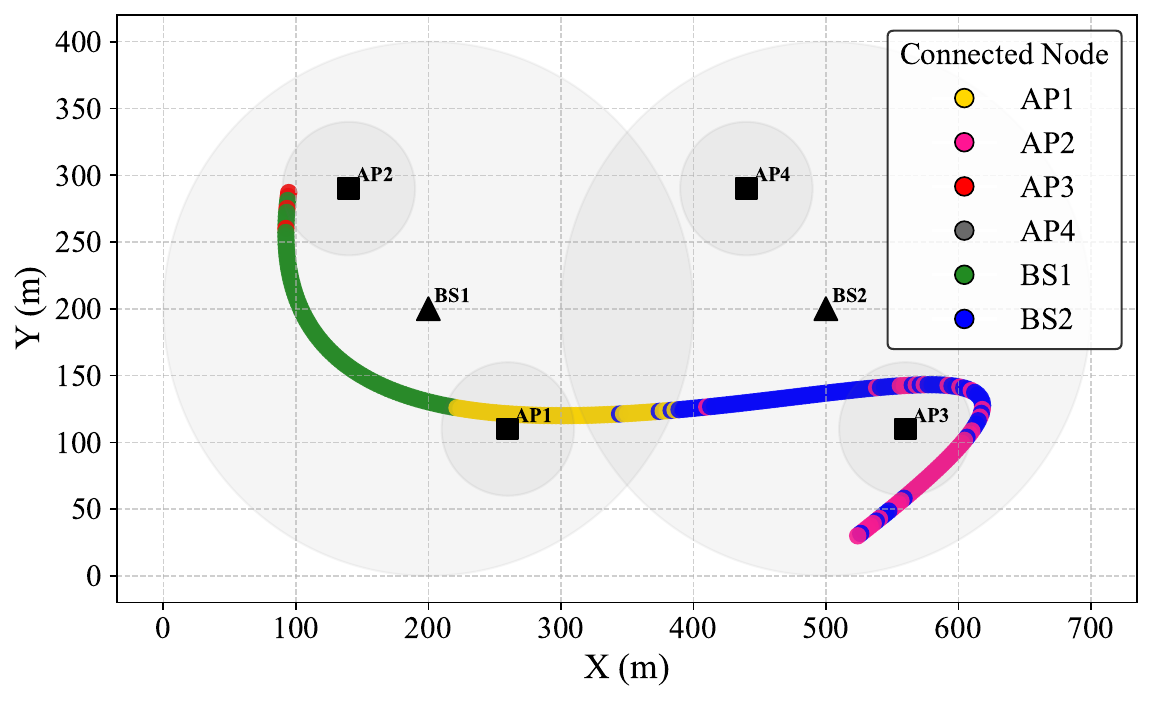} 
\subcaption{}
\label{fig:subim11}
\end{minipage}
\begin{minipage}[t]{0.33\textwidth}
\centering
\includegraphics[width=\textwidth,height=4.7cm]{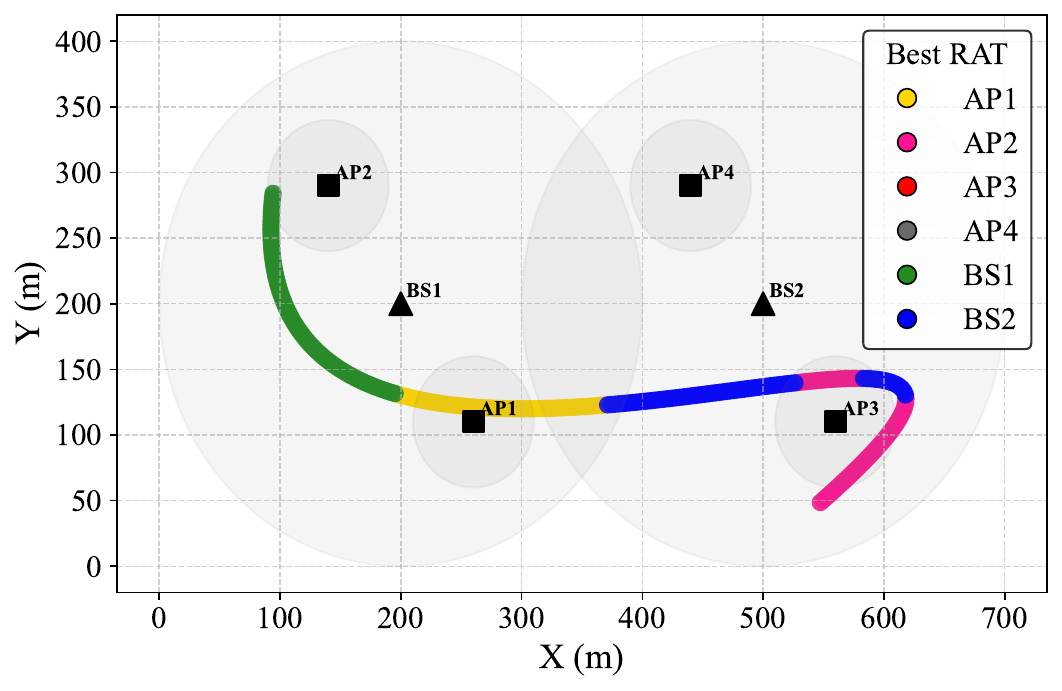} 
\subcaption{}
\label{fig:subim12}
\end{minipage}
\begin{minipage}[t]{0.32\textwidth}
\centering
\includegraphics[width=\textwidth,height=4.7cm]{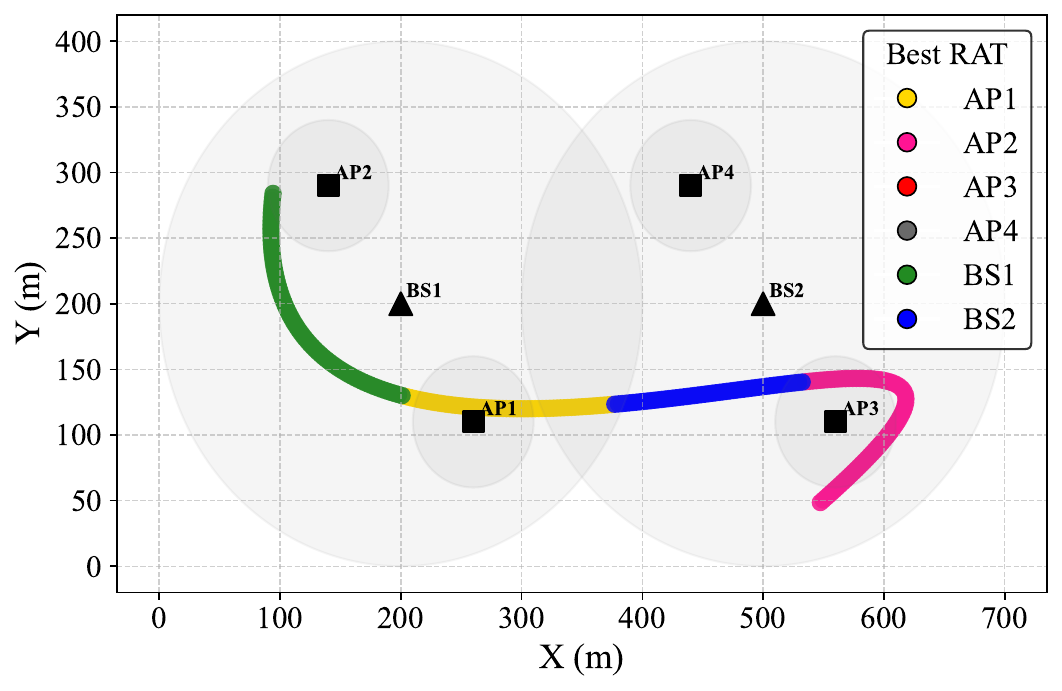}
\subcaption{}
\label{fig:subim13}
\end{minipage}
\caption{The selected serving RAT along the testing UE trajectory, color-coded with different colors per RAT transmitter. (a) Association is based on the actual best-SINR server. (b) Association is selected by the Soft P-CHO decisions. (c) Association is selected by the Hysterisis-enabled P-CHO decisions.}
\label{fig:image8}
\end{figure*}

For this section, to  assess generalization capabilities, we test the handover performance (selected RAT, handover events and stability) of both P-CHO variants, considering a testing UE trajectory that is unseen during the training of SINR predictors. The selected UE trajectory represents a complete path across the two cellular BSs and three APs (see the selected UE path in Fig.~\ref{fig:image8}).

In Fig.~\ref{figure_8}, we report the number of handover occurrences on the testing trajectory as a function of $\delta_{\mathrm{QoS}}$ for both soft and hysteresis-enabled P-CHO variants. Increasing the P-CHO threshold from $0$ dB to $3$ dB reduces the total handovers for both schemes, reflecting more conservative decisions at higher margins and lower handover sensitivity levels. Across all P-CHO thresholds, the hysteresis-enabled scheme consistently yields fewer handovers than the soft P-CHO, demonstrating improved stability and reduced signaling without noticeable loss of connectivity along this path. This means that hysteresis-enabled method adds a small delay to observe more stable and systematic handover gains, without reacting to short and transient advantages of a candidate RAT, which suggests a P-CHO behavior that suppresses ping-pong events.

To visualize the handover behavior of both methods, Fig.~\ref{fig:image8} depicts the selected serving RAT decisions (color-coded) along the testing trajectory, separately for the Soft and Hysteresis-enabled P-CHO (for $N=3$ steps of hysteresis). Both schemes are set at a P-CHO threshold of $\delta_{QoS}=2.5$ dB. The actual best-SINR servers are shown in Fig.~\ref{fig:image8}(a), where it is evident that the best-SINR node frequently varies along the UE path due to dynamic downlink conditions and mobility. In Fig.~\ref{fig:image8}(b), we show the UE-node association derived by directly following the condition in \eqref{eq:soft}, while Fig.~\ref{fig:image8}(c) shows the best RAT selected by following \eqref{eq:hard}. Evidently, the Soft P-CHO exhibits short dwell segments and five switches in the overlap regions between RATs (e.g., near the transitions AP3$\rightarrow$BS2 and BS2$\rightarrow$AP3). This confirms its responsiveness but also its susceptibility to small transient SINR advantages that do not persist, reflecting classic ping-pong behavior when $\Delta\mathrm{QoS}$ varies around the threshold. Second, the Hysteresis-enabled P-CHO maintains longer contiguous segments on the selected RAT and suppresses rapid back-and-forth changes around cell borders. Although this sometimes delays a switch, it eliminates non-beneficial changes caused by short fluctuations or prediction noise. Third, both policies converge to the same serving RAT over most of the trajectory, indicating that prediction-conditioned triggers preserve the correct association pattern. As a result, the hysteresis simply enforces temporal consistency before execution, leading fewer and more decisive handovers. These are translated into reduced signaling and a lower risk of service interruptions, at a small cost in reactivity that can be tuned via $N$ and $\delta_{\mathrm{QoS}}$.

Overall, these results illustrate a clear trade-off: Soft P-CHO maximizes reactivity to predicted gains, while Hysteresis-enabled P-CHO maximizes stability by filtering out short-term advantages. Selecting $(\delta_{\mathrm{QoS}},N)$ provides a practical way to tailor the mobility policy to application needs (e.g., latency-sensitive flows may prefer smaller $N$, whereas throughput-stability or control-plane efficiency may prefer larger $N$).

\section{Conclusions and Future Extensions}

\subsection{Summary and Conclusions}

This paper introduced an ML-assisted conditional and dynamic handover framework for 6G multi-RAT deployments. The proposed P-CHO scheme integrates short-horizon and model-driven signal quality forecasts for all multi-RAT nodes to make proactive decisions. We designed RAT-aware predictors, including a BiLSTM for cellular BSs and a lightweight LSTM for WiFi APs, and integrated them into a generalized P-CHO workflow orchestrated by a RAT Steering Controller. The presented architecture standardizes multiple handover stages, including data collection, parallel model inference, and prediction-conditioned decision logic with hysteresis/dwell time guards, thereby separating preparation from execution in a CHO-compliant manner. Using a Python-based system/link-level simulator for a representative hybrid cellular/WiFi HetNet, we quantified the impact of several system and model parameters. Results show that (i) appropriate prediction windowing improves accuracy, (ii) excessive UE trajectory overlap produces high system stochasticity making signal quality predictions less accurate, and (iii) direct multi-step prediction for long-term forecasts is beneficial compared with recursive inference. Compared against ARIMA and XGBoost baselines, the LSTM models deliver lower RMSE and degrade more gracefully as data complexity grows. Finally, the hysteresis-enabled P-CHO variant consistently reduces unnecessary handovers relative to the soft P-CHO method, improving service stability.

\subsection{Potential Extensions}

Several directions can extend this work. First, the proposed multi-RAT system can be integrated in an Open Radio Access Network (O-RAN) environment by mapping the RAT Steering Controller to the O-RAN Controllers and implementing the signal quality predictors and P-CHO rules as O-RAN xApps. Second, online/federated learning \cite{giannopoulos2024fedship} can be incorporated to adapt to non-stationarity and privacy constraints. In this setup, local signal quality predictors from distributed multi-RAT nodes can be trained in a federated learning manner, without requiring data transfers toward a centralized Controller. Another extension is to jointly optimize prediction and policy via multi-objective control (QoS, energy per bit, HO latency), multi-step planning, or multi-agent RL for multi-UE coordination and admission control. Evaluation of long-term (multi-step) prediction horizons can also be performed for different UE mobility classes. Finally, to ensure low-latency inference of the proposed predictors at the edge, the adoption of model compression, quantization or distillation techniques may be used to reduce model size, while maintaining adequate predictive accuracy.

\section*{Acknowledgment}

The authors warmly thank the partners of UNITY-6G and 6G-Cloud projects for their contribution in the architectural aspects of this article.

\bibliographystyle{IEEEtran}

\end{document}